# Dynamic Non-Bayesian Decision Making

**Dov Monderer** DOV@IE.TECHNION.AC.IL
**Moshe Tennenholtz** MOSHET@IE.TECHNION.AC.IL
*Industrial Engineering and Management*
*Technion – Israel Institute of Technology*
*Haifa 32000, Israel*

## Abstract

The model of a non-Bayesian agent who faces a repeated game with incomplete information against Nature is an appropriate tool for modeling general agent-environment interactions. In such a model the environment state (controlled by Nature) may change arbitrarily, and the feedback/reward function is initially unknown. The agent is not Bayesian, that is he does not form a prior probability neither on the state selection strategy of Nature, nor on his reward function. A policy for the agent is a function which assigns an action to every history of observations and actions. Two basic feedback structures are considered. In one of them – the perfect monitoring case – the agent is able to observe the previous environment state as part of his feedback, while in the other – the imperfect monitoring case – all that is available to the agent is the reward obtained. Both of these settings refer to partially observable processes, where the current environment state is unknown. Our main result refers to the competitive ratio criterion in the perfect monitoring case. We prove the existence of an efficient stochastic policy that ensures that the competitive ratio is obtained at almost all stages with an arbitrarily high probability, where efficiency is measured in terms of rate of convergence. It is further shown that such an optimal policy does not exist in the imperfect monitoring case. Moreover, it is proved that in the perfect monitoring case there does not exist a deterministic policy that satisfies our long run optimality criterion. In addition, we discuss the maxmin criterion and prove that a deterministic efficient optimal strategy does exist in the imperfect monitoring case under this criterion. Finally we show that our approach to long-run optimality can be viewed as qualitative, which distinguishes it from previous work in this area.

## 1. Introduction

Decision making is a central task of artificial agents (Russell & Norvig, 1995; Wellman, 1985; Wellman & Doyle, 1992). At each point in time, an agent needs to select among several actions. This may be a simple decision, which takes place only once, or a more complicated decision where a series of simple decisions has to be made. The question of "what should the right actions be" is the basic issue discussed in both of these settings, and is of fundamental importance to the design of artificial agents.

A static decision-making context (problem) for an artificial agent consists of a set of actions that the agent may perform, a set of possible environment states, and a utility/reward function which determines the feedback for the agent when it performs a particular action in a particular state. Such a problem is best represented by a matrix with columns indexed by the states, rows indexed by the actions and the rewards as entries. When the reward function is not known to the agent we say that the agent has *payoff uncertainty* and we refer to the problem as a *problem with incomplete information*(Fudenberg & Tirole, 1991). When modeling a problem with incomplete information one must also describe the underlying assumptions on the knowledge of the agent about the reward function. For example, the agent may know bounds on his rewards, or he may know (or partially know) an underlying probabilistic structure[1]. In a dynamic (multistage) decision-making setup the agent faces static decision problems over stages. At each stage the agent selects an action to be performed and the environment selects a state. The history of actions and states determines the immediate reward as well as the next one-shot decision problem. The history of actions and states also determines the next selected state. Work on reinforcement learning in artificial intelligence (Kaelbling, Littman, & Moore, 1996) has adopted the view of an agent operating in a probabilistic Bayesian setting, where the agent's last action and the last state determine the next environment state based on a given probability distribution. Naturally, the learner may not be a-priori familiar with this probability distribution, but the existence of the underlying probabilistic model is a key issue in the system's modeling. However, this assumption is not an ultimate one. In particular, much work in other areas in AI and in economics have dealt with non-probabilistic settings in which the environment changes in an unpredictable manner[2]. When the agent does not know the influence of his choices on the selection of the next state (i.e., he is not certain about the environment strategy), we say that the agent has *strategic uncertainty.*

In this paper we use a general model for the representation of agent-environment interactions in which the agent has both payoff and strategic uncertainty. We deal with a non-Bayesian agent who faces a repeated game with incomplete information against Nature.

In a repeated game against Nature the agent faces the same static decision problem at each stage while the environment state is taken to be an action chosen by his opponents. The decision problem is called a game to stress the fact that the agent's action and the state are independently chosen. The fact that the game is repeated refers to the fact that the set of actions, the set of possible states, and the one shot utility function do not vary with time[3]. As we said, we consider an agent that has both payoff uncertainty and strategic uncertainty. That is, he is a-priori ignorant about the utility function (i.e., the game is of incomplete information) as well as about the state selection strategy of Nature. The agent is non-Bayesian in the sense that he does not assume any probabilistic model concerning nature's strategy and in the sense that he does not assume any probabilistic model concerning the reward function, though he may assume lower and upper bounds[4]. We consider two examples to illustrate the above-mentioned notions and model. Consider

1. For example, the agent may know a probability distribution on a set of reward functions, he may assume that such a probability exists without any assumption on its structure, or he may have partial information about this distribution but be ignorant about some of its parameters (e.g., he may believe that the reward function is drawn according to a normal distribution with an unknown covariance matrix).
2. There are many intermediate cases where it is assumed that the changes are probabilistic with a non-Markovian structure.
3. In the most general setup, those sets may vary with time. No useful analysis can be done in a model where those changes are completely arbitrary.
4. Repeated games with complete information, or more generally, multistage games and stochastic games have been extensively studied in game theory and economics. A very partial list includes: (Shapley, 1953; Blackwell, 1956; Luce & Raiffa, 1957), and more recently (Fudenberg & Tirole, 1991; Mertens, Sorin, & Zamir, 1995), and the evolving literature on learning (e.g., Fudenberg & Levine 1997). The incomplete information setup in which the player is ignorant about the game being played was inspired an investor, $I$, who is investing daily in a certain index of the stock market. His daily profits depends on his action (selling or buying in a certain amount) and on the environment state – the percentage change in the price of the index. This investor has complete information about the reward function because he knows the reward which is realized in a particular investment and a particular change, but he has strategic uncertainty about the changes in the index price. So, he is playing a repeated game with complete information against Nature with strategic uncertainty.

Consider another investor, $I1$, who invests in a particular mutual fund. This fund invests in the stock market with a strategy which is not known to the investor. Assume that each state represents the vector of percentage changes in the stocks, then the investor does not know his reward function. For example, he cannot say in advance what would be his profit if he would buy one unit of this fund and all stock prices increase in 1 percent. Thus, $I1$ plays a repeated game with incomplete information. If in addition $I1$ does not attempt to construct a probabilistic model concerning his reward function or market behavior, then he is non-Bayesian and our analysis may apply to him. For another example, assume that Bob has to decide on each evening whether to prepare tea or coffee for his wife before she gets home. His wife wishes to drink either tea or coffee and he wishes to have it ready for her. The reaction of Bob's wife to tea or coffee may depend on her state that day, which can not be predicted based on the history of actions and states in previous days. As Bob has just got married he cannot tell what reward he will get if his wife is happy and he makes her a cup of tea. Of course he may eventually know it, but his decisions during this learning period are precisely the subject of this paper.

As an example for the generality of the above-mentioned setting, consider the model of Markov decision processes with complete or incomplete information. In a Markov decision process an agent's action in a given state determines (in a probabilistic fashion) the next state to be obtained. That is, the agent has a structural assumption on the state selection strategy. A repeated game against Nature without added assumptions captures the fact that the transition from state to state may depend on the history in an arbitrary way.

When the agent performs an action $a_t$ in state $s_t$, part of his feedback would be $u(a_t, s_t)$, where $u$ is the reward function. We distinguish between two basic feedback structures. In one of them – the *perfect monitoring* case – the agent is able to observe the previous environment state as part of his feedback, while on the other – the *imperfect monitoring* case – all that is available to the agent is the reward obtained[5]. Notice that in both of these feedback structures, the current state is not observed by the agent when he is called to select an action[6]. Both investors $I$ and $I1$ face a repeated game with perfect monitoring because the percentage changes become public knowledge after each iteration.

In the other example, when Bob has to make his decision, if the situation is of imperfect monitoring, Bob would be only able to observe the reward for his behavior (e.g., whether by (Harsanyi, 1967). See Aumann and Maschler (1995) for a comprehensive survey. Most of the above literature deals with (partially) Bayesian agents. Some of the rare exceptions are cited in Section 6.

5. Notice that the former assumption is very popular in the related game theory literature (Aumann & Maschler, 1995). Many other intermediate monitoring structures may be interesting as well.

6. Such is also the case in the evolving literature on the problem of controlling partially observable Markov decision processes (Lovejoy, 1991; Cassandra, Kaelbling, & Littman, 1994; Monahan, 1982). In contrast, Q-learning theory (Watkins, 1989; Watkins & Dayan, 1992; Sutton, 1992) does assume a knowledge of the current state.

she says "thanks", "that's terrible", "this is exactly what I wanted", etc.). In the perfect monitoring case, Bob will be able to observe his wife's state (e.g., whether she came home happy, sad, nervous, etc.) in addition to his reward. In both cases Bob (like the investors) is not able to observe his wife's state before making his decision in a particular day.

Consider the case of a one stage game against Nature, in which the utility function is known, but the agent cannot observe the current environment state when selecting his action. How should the agent choose his action? Work on decision making under uncertainty has suggested several approaches (Savage, 1972; Milnor, 1954; Luce & Raiffa, 1957; Kreps, 1988). One of these approaches is the maxmin (safety-level) approach. According to this approach the agent would choose an action that maximizes his worst case payoff. Another approach is the competitive ratio approach (or its additive variant, termed the minmax regret decision criterion (Milnor, 1954). According to this approach an agent would choose an action that minimizes the worst case ratio between the payoff he could have obtained had he known the environment state to the payoff he would actually obtain.[7] Returning back to our example, if Bob would have known the actual state of his wife, he could choose an action that maximizes his payoff. Since he has no hint about her state, he can go ahead and choose the action that minimizes his competitive ratio. For example, if this action leads to a competitive ratio of two, then Bob can guarantee himself that the payoff he would get is at most half the payoff he could have gotten had he known the actual state of his wife.

Given a repeated game with incomplete information against Nature, the agent would not be able to choose his one stage optimal action (with respect to the competitive ratio or maxmin value criteria) at each stage, since the utility function is initially unknown. So, if Bob does not initially know the reward he would receive for his actions as a function of his wife's state, then he will not be able to simply choose an action that guarantees the best competitive ratio. This calls for a precise definition of a long-run optimality criterion. In this paper we are mainly concerned with policies (strategies) guaranteeing that the optimal competitive ratio (or the maxmin value) is obtained in *most* stages. We are interested in particular in efficient policies, where efficiency is measured in terms of rate of convergence. Hence in Bob's case, we are interested in a policy that if adopted by Bob would guarantee him on almost any day, with high probability, at least the payoff guaranteed by an action leading to the competitive ratio. Moreover, Bob will not have to wait much before he will start getting this type of satisfactory behavior.

In Section 2 we define the above mentioned setting and introduce some preliminaries. In Sections 3 and 4 we discuss the long-run competitive ratio criterion: In Section 3 we show that even in the perfect monitoring case, a deterministic optimal policy does not exist. However, we show that there exists an efficient stochastic policy which ensures that the long-run competitive ratio criterion holds with a high probability. In Section 4 we show that such stochastic policies do not exist in the imperfect monitoring case. In Section 5 we prove that for both the perfect and imperfect monitoring cases there exists a deterministic efficient optimal policy for the long-run maxmin criterion. In Section 6 we compare our notions of long-run optimality to other criteria appearing in some of the related literature. In particular, we show that our approach to long-run optimality can be interpreted as

7. The competitive ratio decision criterion has been found to be most useful in settings such as on-line algorithms (e.g., Papadimitriou & Yanakakis, 1989).

qualitative, which distinguishes it from previous work in this area. We also discuss some of the connections of our work with work in reinforcement learning.

## 2. Preliminaries

A (one-shot) *decision problem* (with payoff certainty and strategic uncertainty) is a 3-tuple $D =< A, S, u >$, where $A$ and $S$ are finite sets and $u$ is a real-valued function defined on $A \times S$ with $u(a, s) > 0$ for every $(a, s) \in A \times S$. Elements of $A$ are called *actions* and those of $S$ are called *states*. $u$ is called the *utility function*. The interpretation of the numerical values $u(a, s)$ is context-dependent[8]. Let $n_A$ denote the number of actions in $A$, let $n_S$ denote the number of states in $S$ and let $n = \max(n_A, n_S)$.

The above-mentioned setting is a classical static setting for decision making, where there is uncertainty about the actual state of nature (Luce & Raiffa, 1957). In this paper we deal with a dynamic setup, in which the agent faces the decision problem $D$, without knowing the utility function $u$, over an infinite number of stages, $t = 1, 2, \ldots$. As we have explained in the introduction, this setting enables us to capture general dynamic non-Bayesian decision-making contexts, where the environment may change its behavior in an arbitrary and unpredictable fashion. As mentioned in the introduction, this is best captured by means of a repeated game against Nature. The state of the environment at each point plays the role of an action taken by Nature in the corresponding game. The agent knows the sets $A$ and $S$, but he does not know the payoff function $u$.[9] A *dynamic decision problem* (with payoff uncertainty and strategic uncertainty) is therefore represented for the agent by a pair $DD =< A, S >$ of finite sets[10]. At each stage $t$, Nature chooses a state $s_t \in S$. The agent, who does not know the chosen state, chooses $a_t \in A$, and receives $u(a_t, s_t)$. We distinguish between two informational structures. In the *perfect monitoring* case, the state $s_t$ is revealed to the agent alongside the payoff $u(a_t, s_t)$. In the *imperfect monitoring* case, the states are not revealed to the agent. A generic history available to the agent at stage $t+1$ is denoted by $h_t$. In the perfect monitoring case, $h_t \in H_t^p = (A \times S \times R_+)^t$, where $R_+$ denotes the set of positive real numbers. In the imperfect monitoring case, $h_t \in H_t^{imp} = (A \times R_+)^t$. In the particular case $t = 0$ we assume that $H_0^p = H_0^{imp} = \{e\}$ is a singleton containing the empty history $e$. Let $H^p = \cup_{t=0}^{\infty} H_t^p$ and let $H^{imp} = \cup_{t=0}^{\infty} H_t^{imp}$. The symbol $H$ will be used for both $H^p$ and $H^{imp}$. A *strategy*[11] for the agent in a dynamic decision problem is a function $F : H \to \Delta(A)$ , where $\Delta(A)$ denotes the set of probability measures over $A$. That is, for every $h_t \in H$, $F(h_t) : A \to [0, 1]$ and $\sum_{a \in A} F(h_t)(a) = 1$. In other words, if the agent observes the history $h_t$ then he chooses $a_{t+1}$ by randomizing amongst his actions, with the probability $F(h_t)(a)$ assigned to the action $a$. A strategy $F$ is called *pure* if $F(h_t)$ is a probability measure concentrated on a singleton for every $t \geq 0$.

In Sections 2–4 the strategy recommended to the agent is chosen according to a "long-run" decision criterion which is induced by the *competitive ratio* one-stage decision criterion.

8. See the discussion at Section 6.
9. All the results of this paper remain unchanged if the agent does not initially know the set $S$, but rather an upper bound on $n_S$.
10. Notice that there is no need to include an explicit transition function in this representation. This is due to the fact that in the non-Bayesian setup every transition is possible.
11. Strategy is a decision theoretic concept. It coincides with the term *policy* used in the control theory literature, and with the term *protocol* used in the distributed systems literature.

The competitive ratio decision criterion, that is described below, may be used by an agent who faces the decision problem only once, and who knows the payoff function $u$ as well as the sets $A$ and $S$. There are other "reasonable" decision criteria that could be used instead. One of them is the *maxmin* decision criterion to be discussed in Section 5, while another is the minmax regret decision criterion (Luce & Raiffa, 1957; Milnor, 1954). The latter is a simple variant of the competitive ratio (and can be treated similarly), and therefore will not be treated explicitly in this paper.

For every $s \in S$ let $M(s)$ be the maximal payoff the agent can get when the state is $s$. That is

$$M(s) = \max_{a \in A} u(a, s).$$

For every $a \in A$ and $s \in S$ define

$$c(a, s) = \frac{M(s)}{u(a, s)}.$$

Denote $c(a) = \max_{s \in S} c(a, s)$, and let

$$CR = \min_{a \in A} c(a) = \min_{a \in A} \left( \max_{s \in S} c(a, s) \right).$$

$CR$ is called the *competitive ratio* of $D =< A, S, u >$. Any action $a$ for which $CR = c(a)$ is called a *competitive ratio action*, or in short a $CR$ action. An agent which chooses a $CR$ action guarantees receiving at least $\frac{1}{CR}$ fraction from what it could have gotten, had it known the state $s$. That is, $u(a, s) \geq \frac{1}{CR} M(s)$ for every $s \in S$. This agent cannot guarantee a bigger fraction.

In the long-run decision problem, a non-Bayesian agent does not form a prior probability on the way Nature is choosing the states. Nature may choose a fixed sequence of states or, more generally, use a probabilistic strategy $G$, where $G : Q \rightarrow \Delta(S)$, and $Q = \cup_{t=0}^{\infty} Q_t = \cup_{t=0}^{\infty} (A \times S)^t$. Nature can be viewed as a second player that knows the reward function. Its strategy may of course refer to the whole history of actions and states until a given point and may depend on the payoff function.

A payoff function $u$ and a pair of probabilistic strategies $F, G$, where $G$ can depend on $u$, generate a probability measure $\mu = \mu_{F,G,u}$ over the set of infinite histories $Q_\infty = (A \times S)_\infty$ endowed with the natural measurable structure. For an event $B \subseteq Q_\infty$ we will denote the probability of $B$ according to $\mu$ by $\mu(B)$ or by $Prob_\mu(B)$. More precisely, the probability measure $\mu$ is uniquely defined by its values for finite cylinder sets: Let $A_t : Q_\infty \rightarrow A$ and $S_t : Q_\infty \rightarrow S$ be the coordinate random variables which contain the values of the actions and states selected by the agent and the environment in stage $t$ (respectively). That is, $A_t(h) = a_t$ and $S_t(h) = s_t$ for every $h = ((a_1, s_1), (a_2, s_2), \ldots)$ in $Q_\infty$. Then for every $T \geq 1$ and for every $((a_1, s_1), \ldots, (a_T, s_T)) \in Q_T$,

$$Prob_\mu \left( (A_t, S_t) = (a_t, s_t) \quad \text{for all } 1 \leq t \leq T \right) = \prod_{t=1}^{T} F(\varphi_{t-1})(a_t) G(\psi_{t-1})(s_t),$$

where $\psi_0$ and $\varphi_0$ are the empty histories, and for $2 \leq t \leq T$ we have

$$\psi_{t-1} = \left( (a_1, s_1), \ldots, (a_{t-1}, s_{t-1}) \right),$$

while the definition of $\varphi_{t-1}$ depends on the monitoring structure. In the perfect monitoring case,

$$\varphi_{t-1} = ((a_1, s_1, u(a_1, s_1)), \ldots, (a_{t-1}, s_{t-1}, u(a_{t-1}, s_{t-1}))) ,$$

and in the imperfect monitoring case

$$\varphi_{t-1} = ((a_1, u(a_1, s_1)), \ldots, (a_{t-1}, u(a_{t-1}, s_{t-1}))) .$$

We now define some auxiliary additional random variables on $Q_\infty$.

Let $X_t = 1$ if $c(A_t, S_t) \leq CR$ and $X_t = 0$ otherwise, and let $N_T = \sum_{t=1}^{T} X_t$.[12] Let $\delta > 0$. A strategy $F$ is *$\delta$-optimal* if there exists an integer $K$ such that for every payoff function $u$ and every Nature's strategy $G$

$$Prob_\mu(N_T \geq (1-\delta)T \quad \text{for every } T \geq K) \geq 1 - \delta \tag{1}$$

where $\mu = \mu_{F,G,u}$. A strategy $F$ is *optimal* if it is $\delta$-optimal for all $\delta > 0$.

Roughly speaking, $N_T$ measures the number of stages in which the competitive ratio (or a better value) has been obtained in the first $T$ iterations. In an $\delta$-optimal strategy there exists a number $K$, such that if the system runs for $T \geq K$ iterations we can get with high probability that $N_T$ is close to 1 (i.e., almost all iterations are good ones). In an optimal strategy we guarantee that we can get as close as we wish to the situation where all iterations are good ones, with a probability that is as high as we wish. Notice that we require that the above-mentioned useful property will hold for every payoff function and for every strategy of Nature. This strong requirement is a consequence of the non-Bayesian setup; since we do not have any clue about the reward function or about the strategy selected by Nature (and this strategy may yield arbitrary sequences of states to be reached), the best policy would be to insist on good behavior against any behavior adopted by Nature. Notice however that two other relaxations are introduced here; we require successful behavior in most stages, and that the whole process would be successful only with some (very high) probability.

The major objective is to find a policy that will enable (1) to hold for every dynamic decision problem and every Nature strategy. Moreover, we wish (1) to hold for small enough $K$. If $K$ is small then our agent can benefit from obtaining the desired behavior already in an early stage. [13] This will be the subject of the following section. We complete this section with a useful technical observation. We show that a strategy $F$ is $\delta$-optimal if it satisfies the optimality criterion (1) for every reward function and for every stationary strategy of nature, where a stationary strategy is defined by a sequence of states $z = (s_t)_{t=1}^{\infty}$. In this strategy Nature chooses $s_t$ at stage $t$, independent of the history. Indeed, assume that $F$ is a strategy for which (1) holds for every reward function and for every stationary strategy of Nature, then we show that $F$ is $\delta$-optimal.

Given any payoff function $u$ and any strategy $G$, $\delta$-optimality with respect to stationary strategies implies that for $\mu = \mu_{F,G,u}$,

$$Prob_\mu(N_T \geq (1-\delta)T \quad \text{for every } T \geq K)|S_1, S_2, \ldots) \geq 1 - \delta,$$

12. Note that the function $c(a, s)$ depends on the payoff function $u$ and therefore so do the random variables $X_t$ and $N_t$.

13. The interested reader may wish to think of our long-run optimality criteria in view of the original work on PAC learning (Valiant, 1984). In our setting, as in PAC learning, we wish to obtain desired behavior, in most situations, with high probability, and relatively fast.

with probability one. Therefore

$$Prob_{\mu}(N_T \geq (1-\delta)T \quad \text{for every } T \geq K) \geq 1-\delta.$$

Roughly speaking, the above captures the fact that in our non-Bayesian setting we need to present a strategy that will be good for any sequence of states chosen by Nature, regardless of the way in which it has been chosen.

## 3. Perfect Monitoring

In this section we present our main result. This result refers to the case of perfect monitoring, and shows the existence of a $\delta$-optimal strategy in this case. It also guarantees that the desired behavior will be obtained after polynomially many stages. Our result is constructive. We first present the rough idea of the strategy employed in our proof. If the utility function was known to the agent then he could use the competitive ratio action. Since the utility function is initially unknown then the agent will use a greedy strategy, where he selects an action that is optimal as far as the competitive ratio is concerned, according to the agent's knowledge at the given point. However, the agent will try from time to time to sample a random action.[14] Our strategy chooses a right tradeoff between these exploration and exploitation phases in order to yield the desired result. Some careful analysis is needed in order to prove the optimality of the related strategy, and the fact it yields the desired result after polynomially many stages.

We now introduce our main theorem.

**Theorem 3.1:** *Let $DD =< A, S >$ be a dynamic decision problem with perfect monitoring. Then for every $\delta > 0$ there exists a $\delta$-optimal strategy. Moreover, the $\delta$-optimal strategy can be chosen to be efficient in the sense that $K$ (in (1)) can be taken to be polynomial in $max(n, \frac{1}{\delta})$.*

**Proof:** Recall that $n_A$ and $n_S$ denote the number of actions and states respectively, and that $n = \max(n_A, n_S)$. In this proof we assume for simplicity that $n = n_A = n_S$. Only slight modifications are required for removing this assumption. Without loss of generality, $\delta < 1$. We define a strategy $F$ as follows: Let $M = \frac{8}{\delta}$. That is,

$$\frac{1}{M} = \frac{\delta}{8}.$$

At each stage $T \geq 1$ we will construct matrices $U_T^F, C_T^F$ and a subset of the actions $W_T$ in the following way: Define $U_1^F(a,s) = *$ for each $a, s$. At each stage $T > 1$, if $a_{T-1}$ has been performed in stage $T-1$, and $s_{T-1}$ has been observed, then update $U$ by replacing the $*$ in the $(a_{T-1}, s_{T-1})$ entry with $u(a_{T-1}, s_{T-1})$. At each stage $T \geq 1$, if $U_T^F(a,s) = *$, define $C_T^F(a,s) = 1$. If $U_T^F(a,s) \neq *$, $C_T^F(a,s) = \max_{\{b: U_T^F(b,s) \neq *\}} \frac{U_T^F(b,s)}{U_T^F(a,s)}$. Finally $W_T$ is the set of all $a \in A$ at which $\min_{b \in A} \max_{s \in S} C_T^F(b,s)$ is obtained. We refer to elements in $W_T$ as the *temporarily good* actions at stage $T$. Let $(Z_t)_{t \geq 1}$ be a sequence of i.i.d. $\{0,1\}$ random

14. We use a uniform probability distribution to select among actions in the exploration phase. Our result can be obtained with different exploration techniques as well.

variables with $Prob(Z_t = 1) = 1 - \frac{1}{M}$. This sequence is generated as part of the strategy, independent of the observed history. That is at each stage, before choosing an action, the agent flips a coin, independently of his past observations. At each stage $t$ the agent observes $Z_t$. If $Z_t = 1$, the agent chooses an action from $W_t$ by randomizing with equal probabilities. If $Z_t = 0$ the agent randomizes with equal probabilities amongst the actions in $A$. This complete the description of the strategy $F$. Let $u$ be a given payoff function, and let $(s_t)_{t=1}^{\infty}$ be a given sequence of states. We proceed to show that (1) holds with $K$ being the upper integer value of $\alpha = \max(\alpha_1 + 2, \alpha_2 + 2)$, where

$$\alpha_1 = \frac{128}{\delta^2} \ln\left(\frac{256}{\delta^3}\right) \quad \text{and} \quad \alpha_2 = \frac{n^2(n\frac{8}{\delta} + 1)\ln\left(\frac{2n^2}{\delta}\right) + 1}{\frac{3}{4}\delta}.$$

Recall that $X_t = 1$ if $c(A_t, s_t) \leq CR$ and $X_t = 0$ otherwise, and that $N_T = \sum_{t=1}^{T} X_t$. By a slight change of notation, we denote by $P_\mu = Prob_\mu$ the probability measure induced by $F$ , $u$ and the sequence of states on $(A \times S \times \{0,1\})_\infty$ (where $\{0,1\}$ corresponds to the $Z_t$ values).

Let $\varepsilon = \frac{\delta}{8}$. Define

$$B_K = \left\{ \sum_{t=1}^{T} Z_t \geq \left(1 - \frac{1}{M} - \varepsilon\right) T \quad \text{for all } T \geq K \right\}.$$

Roughly speaking, $B_K$ captures the cases where temporarily good actions are selected in most stages.

By (Chernoff, 1952) (see also (Alon, Spencer, & Erdos, 1992)), for every $T$,

$$P_\mu\left(\sum_{t=1}^{T} Z_t < (1 - \frac{1}{M} - \varepsilon)T\right) \leq e^{-\frac{\varepsilon^2 T}{2}}.$$

Recall that given a set $S$, $\overline{S}$ denotes the complement of $S$.

Hence,

$$P_\mu(\overline{B_K}) \leq \sum_{T=k}^{\infty} P_\mu\left(\sum_{t=1}^{T} Z_t < (1 - \frac{1}{M} - \varepsilon)T\right) \leq \sum_{T=K}^{\infty} e^{-\frac{\varepsilon^2 T}{2}}.$$

Therefore

$$P_\mu(\overline{B_K}) \leq \int_{k-1}^{\infty} e^{-\frac{\varepsilon^2 T}{2}} dT = \frac{2}{\varepsilon^2} e^{-\frac{\varepsilon^2 (K-1)}{2}}.$$

Since $K > \alpha_1$,

$$P_\mu(\overline{B_K}) < \frac{\delta}{2} \tag{2}$$

Define:

$$L_K = \{N_T \geq (1 - \delta)T \quad \text{for every } T \geq K\}.$$

Roughly speaking, $L_K$ captures the cases where competitive ratio actions (or better actions in this regard) are selected in most stages.

In order to prove that $F$ is $\delta$-optimal (i.e., that (1) is satisfied), we have to prove that

$$P_\mu(\overline{L_K}) < \delta \tag{3}$$

By (2) it suffices to prove that

$$P_\mu(\overline{L_K}|B_K) \leq \frac{\delta}{2} \tag{4}$$

We now define for every $t \geq 1$, $s \in S$ and $a \in A$ six auxiliary random variables, $Y_t$, $R_t$, $Y_t^s$, $R_t^s$, $Y_t^{s,a}$, $R_t^{s,a}$. Let $Y_t = 1$ whenever $Z_t = 1$ and $X_t = 0$, and $Y_t = 0$ otherwise. Let

$$R_T = \sum_{t=1}^{T} Y_t.$$

For every $s \in S$ let $Y_t^s = 1$ whenever $Y_t = 1$ and $s_t = s$, and $Y_t^s = 0$ otherwise. Let

$$R_T^s = \sum_{t=1}^{T} Y_t^s.$$

For every $s \in S$ and for every $a \in A$, let $Y_t^{s,a} = 1$ whenever $Y_t^s = 1$ and $A_t = a$, and $Y_t^{s,a} = 0$ otherwise. Let

$$R_T^{s,a} = \sum_{t=1}^{T} Y_t^{s,a}.$$

Let $g$ be the integer value of $\frac{3}{4}\delta K$. We now show that

$$P_\mu(\overline{L_K}|B_K) \leq P_\mu(\exists T \geq K \quad, R_T \geq g|B_K) \tag{5}$$

In order to prove (5) we show that

$$\overline{L_K} \cap B_K \subseteq \{\exists T \geq K \quad, R_T \geq g\} \cap B_K.$$

Indeed, if $w$ is a path in $B_K$ such that for every $T \geq K$ $R_T < g$, then, at $w$, for every $T \geq K$,

$$N_T \geq \sum_{1 \leq t \leq T, Z_t = 1} X_t \geq V_T - \sum_{t=1}^{T} Y_t,$$

where $V_T$ denotes the number of stages $1 \leq t \leq T$ for which $Z_t = 1$. Since $w \in B_K$,

$$N_T \geq (1 - \frac{1}{M} - \varepsilon)T - R_T > (1 - \frac{1}{M} - \varepsilon)T - g$$

for every $T \geq K$. Since $\frac{1}{M} = \varepsilon = \frac{\delta}{8}$ and $g \leq \frac{3}{4}\delta K$, $N_T \geq (1 - \delta)T$ for every $T \geq K$. Hence, $w \in L_K$.

(5) implies that it suffices to prove that

$$P_\mu(\exists T \geq K \quad, R_T \geq g|B_K) \leq \frac{\delta}{2} \tag{6}$$

Therefore it suffices to prove that for every $s \in S$,

$$P_\mu\left(\exists T \geq K, \quad R_T^s \geq \frac{g}{n}|B_K\right) \leq \frac{\delta}{2n}.$$

Hence it suffices to prove that for every $s \in S$ and every $a \in A$,

$$\gamma = P_\mu\left(\exists T \geq K, \quad R_T^{s,a} \geq \frac{g}{n^2}|B_K\right) \leq \frac{\delta}{2n^2} \tag{7}$$

In order to prove (7) , note that if the inequality $R_T^{s,a} \geq \frac{g}{n^2}$ is satisfied at $w$, then $c(a,s) > CR$ and $a$ is nevertheless considered to be a good action in at least $\frac{g}{n^2}$ stages $1 \leq t \leq T$ (w.l.o.g. assume that $\frac{g}{n^2}$ is an integer). Let $b \in A$ satisfy $\frac{u(b,s)}{u(a,s)} > CR$. If $b$ is ever played in a stage $\bar{t}$ with $s_{\bar{t}} = s$, then $a \notin W_t$ for all $t \geq \bar{t}$. Therefore

$$\gamma \leq P\mu\left(\exists T \geq K, \quad b \text{ is not played in the first } \tfrac{g}{n^2} \text{ stages at which } s_t = s|B_K\right).$$

Hence

$$\gamma \leq \left(1 - \frac{1}{nM}\right)^{\frac{g}{n^2}}.$$

As $(1-\frac{1}{x})^{x+1} \leq e^{-x}$ for $x \geq 1$,

$$\gamma \leq e^{-\frac{g}{n^2(nM+1)}} < \frac{\delta}{2n^2}.$$

□

Theorem 3.1 shows that efficient dynamic non-Bayesian decisions may be obtained by an appropriate stochastic policy. Moreover, it shows that $\delta$-optimality can be obtained in time which is a (low degree) polynomial in $max(n, \frac{1}{\delta})$. An interesting question is whether similar results can be obtained by a pure/deterministic strategy. As the following example shows, deterministic strategies do not suffice for that job.

We give an example in which the agent does not have a $\delta$ optimal *pure* strategy.

**Example 1:** Let $A = \{a^1, a^2\}$ and $S = \{s^1, s^2\}$. Assume in negation that the agent has a $\delta$ optimal pure strategy $f$.

Consider the following two decision problems whose rows are indexed by the actions and whose columns are indexed by the states:

$$D_1 = \begin{pmatrix} 1 & 10 \\ 30 & 2 \end{pmatrix}$$

$$D_2 = \begin{pmatrix} 1 & 30 \\ 10 & 2 \end{pmatrix}$$

with the corresponding ratio matrices:

$$C_1 = \begin{pmatrix} 30 & 1 \\ 1 & 5 \end{pmatrix}$$

$$C_2 = \begin{pmatrix} 10 & 1 \\ 1 & 15 \end{pmatrix}$$

Assume in addition that in both cases Nature uses the strategy $g$ , defined as follows: $g(h_t) = s^i$ if $f(h_t) = a^i$, $i = 1, 2$. That is, for every $t$, $(a_t, z_t) = (a^1, s^1)$ or $(a_t, z_t) = (a^2, s^2)$, where $a_t$ and $z_t$ denote the action and state selected at stage $t$, respectively. Let $\delta < 0.25$. Let $N_T^i$ denote $N_T$ for decision problem $i$. Since $f$ is $\delta$-optimal, there exists $K$ such that for every $T \geq K$, $N_T^1 \geq (1-\delta)T$ and $N_T^2 \geq (1-\delta)T$. Note also that the same sequence $((a_t, z_t))_{t\geq 1}$ is generated in both cases. $N_K^1 \geq (1-\delta)K$ implies that $(a^2, s^2)$ is used at more than half of the stages $1, 2, \ldots, K$. On the other hand, $N_K^2 \geq (1-\delta)K$ implies that $(a^1, s^1)$ is used at more than half of the stages $1, 2, \ldots, K$. A contradiction.
□

For analytical completeness, we end this section by proving the existence of an optimal strategy (and not merely a $\delta$-optimal strategy). Such an optimal strategy is obtained by utilizing $\delta_m$-optimal strategies (whose existence was proved in Theorem 3.1) for intervals of stages with sizes that converge to infinity, when $\delta_m \to 0$.

**Corollary 3.2:** *In every dynamic decision problem with perfect monitoring there exists an optimal strategy.*

**Proof:** For $m \geq 1$, let $F_m$ be a $\frac{\delta_m}{2}$-optimal strategy, where $(\delta_m)_{m=1}^{\infty}$ is a decreasing sequence with $\lim_{m\to\infty} \delta_m = 0$. Let $(K_m)_{m=1}^{\infty}$ be an increasing sequence of integers such that for every $m \geq 1$

$$Prob\left(N_T \geq (1-\frac{\delta_m}{2})T \quad \text{for every } T \geq K_m\right) \geq 1 - \frac{\delta_m}{2},$$

and

$$K_{m+1} \geq \ 2\frac{\sum_{j=1}^{m} K_j}{\delta_m}.$$

Let $F$ be the strategy that for $m \geq 1$ utilizes $F_m$ at the stages $K_0 + \ldots + K_{m-1} + 1 \leq t \leq K_0 + \ldots + K_{m-1} + K_m$, where $K_0 = 0$. It can be easily verified that $F$ is optimal.
□

## 4. Imperfect Monitoring

We proceed to give an example for the imperfect monitoring case, where for sufficiently small $\delta > 0$, the agent does not have a $\delta$-optimal strategy.

### Example 2 (Non-existence of $\delta$-optimal strategies in the imperfect monitoring case)

Let $A = \{a^1, a^2\}$, and $S = \{s^1, s^2, s^3\}$. Let $\delta < \delta_0$, where $\delta_0$ is defined at the end of this proof. Assume in negation that there exists a $\delta$-optimal strategy $F$. Consider the following two decision problems whose rows are indexed by the actions and whose columns are indexed by the states:

$$D_1 = \begin{pmatrix} 2a & 2b & 2c \\ a & b & c \end{pmatrix}$$

$$D_2 = \begin{pmatrix} 2a & 2b & 2c \\ b & c & a \end{pmatrix}$$

where $a, b$ and $c$ are positive numbers satisfying: $a > 4b > 16c$. For $i = 1, 2$, let $C_i = (c_i(a,s))_{a \in A, s \in S}$ be the ratio matrices. That is:

$$C_1 = \begin{pmatrix} 1 & 1 & 1 \\ 2 & 2 & 2 \end{pmatrix}$$

$$C_2 = \begin{pmatrix} 1 & 1 & \frac{a}{2c} \\ \frac{2a}{b} & \frac{2b}{c} & 1 \end{pmatrix}$$

Note that $a^1$ is the unique $CR$ action in $D_1$ and $a^2$ is the unique $CR$ action in $D_2$. Assume that Nature uses strategy $G$ which randomizes at each stage with equal probabilities (of $\frac{1}{3}$) amongst all 3 states. Given this strategy of Nature, the agent cannot distinguish between the two decision problems, even if he knows Nature's strategy and he is told that one of them is chosen. This implies that if $\mu_1$ and $\mu_2$ are the probability measures induced by $F$ and $G$ on $(A \times S)_\infty$ in the decision problems $D_1$ and $D_2$ respectively, then for every $i \in \{1, 2\}$, the distribution of the stochastic process $(N_T^i)_{T=1}^\infty$ with respect to $\mu_j$, $j \in \{1, 2\}$, does not depend on $j$. That is, for every $T \geq 1$

$$Prob_{\mu_1}\left(N_t^i \in M_t \quad \text{for all } t \leq T\right) = Prob_{\mu_2}\left(N_t^i \in M_t \quad \text{for all } t \leq T\right), \quad i \in \{1, 2\} \tag{8}$$

for every sequence $(M_t)_{t=1}^T$ with $M_t \subseteq \{0, 1, \ldots, t\}$ for all $1 \leq t \leq T$.

We do not give a complete proof of (8), rather we illustrate it by proving a representing case. The reader can easily derive the complete proof. We show that

$$Prob_{\mu_1}(N_2^1 = 0) = Prob_{\mu_2}(N_2^1 = 0) \tag{9}$$

Indeed, for $j = 1, 2$,

$$Prob_{\mu_j}(N_2^1 = 0) = \frac{1}{3} \sum_{k=1}^{3} F(e)(a^2) F(a_2, u_j(a^2, s^k))(a^2) \tag{10}$$

Let $\pi : \{1, 2, 3\} \to \{1, 2, 3\}$ be defined by $\pi(1) = 3$, $\pi(2) = 1$, and $\pi(3) = 2$. Then

$$u_1(a^2, s^k) = u_2(a^2, s^{\pi(k)})$$

for every $1 \leq k \leq 3$. Therefore (10) implies (9).

As $F$ is $\delta$-optimal, then there exists an integer $K$ such that with a probability of at least $1 - \delta$ with respect to $\mu_1$, and hence with respect to $\mu_2$, $N_T^1 \geq (1 - \delta)T$ for every $T \geq K$. This implies that with a probability of at least $1 - \delta$, $a^1$ is played at least at $1 - \delta$ of the stages up to time $T$, for all $T \geq K$, and in particular for $T = K$. We choose the integer $K$ to be sufficiently large such that according to the Law of Large Numbers, Nature chooses $s^3$ in at least $\frac{1}{3} - \delta$ of the stages up to stage $K$ with a probability of at least $1-\delta$. Let $CR_2$ and $c_t^2$ denote $CR$ and $c_t$ of decision problem 2, respectively. Then

$$\frac{a}{2c} > CR_2 = \max(\frac{2a}{b}, \frac{2b}{c}).$$

Therefore, if $A_t = a^1$, then $C_2(A_t, S_t) \leq CR_2$ if and only if $S_t \neq s^3$. Hence, with a probability of at least $1 - 2\delta$, in at most $\delta + (1-\delta)(\frac{2}{3} + \delta)$ of these stages $c_t^2 \leq CR_2$. Therefore $F$ cannot be $\delta$-optimal, if we choose $\delta_0$ such that $2\delta_0 < 1 - \delta_0$ and

$$\delta_0 + (1-\delta_0)(\frac{2}{3} + \delta_0) < 1 - \delta_0.$$

□

## 5. Safety Level

For the sake of comparison we discuss in this section the safety level (known also as maxmin) criterion. Let $D =< A, S, u >$ be a decision problem. Denote

$$v = \max_a \min_s u(a, s)$$

$v$ is called the *safety level* of the agent (or the maxmin value). Every action $a$ for which $u(a, s) \geq v$ for every $s$ is called a safety level action. We consider now the imperfect monitoring model for the dynamic decision problem $< A, S >$. Every sequence of states $z = (s_t)_{t=1}^{\infty}$ with $s_t \in S$ for every $t \geq 1$ and every pure strategy $f$ of the agent induce a sequence of actions $(a_t)_{t=1}^{\infty}$ and a corresponding sequence of payoffs $(u_t^{z,f})_{t=1}^{\infty}$, where $u_t^{z,f} = u(a_t, s_t)$ for every $t \geq 1$. Let $N_T^{z,f}$ denote the number of stages up to stage $T$ at which the agent's payoff exceeds the safety level $v$. That is,

$$N_T^{z,f} = \#\{1 \leq t \leq T : \quad u_t^{z,f} \geq v\} \tag{11}$$

We say that $f$ is *safety level optimal* if for every decision problem and for every sequence of states,

$$\lim_{T \to \infty} \frac{1}{T} N_T^{z,f} = 1,$$

and the convergence holds uniformly w.r.t. all payoff functions $u$ and all sequences of states in $S$. That is, for every $\delta > 0$ there exists $K = K(\delta)$ such that $N_T^{z,f} \geq (1-\delta)T$ for every $T \geq K$ for every decision problem $< A, S, u >$ and for every sequence of states $z$.

**Proposition 5.1:** *Every dynamic decision problem possesses a safety level optimal strategy in the imperfect monitoring case, and consequently in the perfect monitoring case. Moreover, such an optimal strategy can be chosen to be strongly efficient in the sense that for every sequence of states there exists at most $n_A - 1$ stages at which the agent receives a payoff smaller than his safety level, where $n_A$ denotes the number of actions.*

**Proof:** Let $n = n_A$. Define a strategy $f$ as follows: Play each of the actions in $A$ in the first $n$ stages. For every $T \geq n+1$, and for every history $h = h_{T-1} =$ $((a_1, u_1), (a_2, u_2, \ldots, (a_{T-1}, u_{T-1}))$ we define $f(h) \in A$ as follows: For $a \in A$, let $v_T^h(a) = \min u_t$, where the min ranges over all stages $t \leq T-1$ for which $a_t = a$. Define $f(h) = a_T$, where $a_T$ maximizes $v_T^h(a)$ over $a \in A$. It is obvious that for every sequence of states $z = (s_t)_{t=1}^{\infty}$ there are at most $n-1$ stages at which $u(a_t, s_t) < v$, where $(a_t)_{t=1}^{\infty}$ is the sequence of actions generated by $f$ and the sequence of states. Hence $N_T^{z,f} \geq T - n$, where $N_T^{z,f}$ is defined in (11). Thus for $K(\delta) = \frac{n}{\delta}$, $\frac{1}{T} N_T^{z,f} \geq 1 - \delta$ for every $T \geq K(\delta)$.
□

## 6. Discussion

Note that all the notations established in Section 5, and Proposition 5.1 as well, remain unchanged if we assume that the utility function $u$ takes values in a totally pre-ordered set without any group structure. That is, our approach to decision making is qualitative (or ordinal). This distinguishes our work from previous work on non-Bayesian repeated games, which used the probabilistic safety level criterion as a basic solution concept for the one shot game[15]. These works, including (Blackwell, 1956; Hannan, 1957; Banos, 1968; Megiddo, 1980), and more recently (Auer, Cesa-Bianchi, Freund, & Schapire, 1995; Hart & Mas-Colell, 1997), used several versions of long-run solution concepts, all based on some optimization of the average of the utility values over time. That is, in all of these papers the goal is to find strategies that guarantee that with high probability $\frac{1}{T}\sum_{t=1}^{T} u(A_t, S_t)$ will be close to $v_p$.

Our work is, to the best of our knowledge, the first to introduce an efficient dynamic optimal policy for the basic competitive ratio context. Our study and results in sections 2-4 can be easily adapted to the case of *qualitative* competitive ratio as well. In this approach, the utility function takes values in some totally pre-ordered set $G$ and in addition we assume a *regret function*, $\psi$ that maps $G \times G$ to some pre-ordered set $H$. For $g_1, g_2 \in G$, $\psi(g_1, g_2)$ is the level of regret when the agent receives the utility level $g_1$ rather than $g_2$. Given an action $a$ and a state $s$, the regret function will determine the maximal regret, $c(a, s) \in H$ of the agent when action $a$ is performed in state $s$. That is,

$$c(a, s) = \max \psi(u(a, s), u(b, s)),$$

where $b$ ranges over all actions.

The qualitative regret of action $a$ will be the maximal regret of this action over all states. The optimal qualitative competitive ratio will be obtained by using an action for which the qualitative regret is minimal. Notice that no arithmetic calculations are needed (or make any sense) for this qualitative version. Our results can be adapted to the case of qualitative competitive ratio. For ease of exposition, however, we used the quantitative version of this model, where a numerical utility function represents the regret function.

15. The probabilistic safety value, $v_p$, of the agent in the decision problem $D = < A, S, u >$ is his maxmin value when the max ranges over all mixed actions. That is

$$v_p = max_{q \in \Delta(A)} min_{s \in S} \sum_{a \in A} u(a, s) q(a),$$

where $\Delta(A)$ is the set of all probability distributions $q$ on $A$.

Our work is relevant to research on reinforcement learning in AI. Work in this area, however, has dealt mostly with Bayesian models. This makes our work complementary to this work. We wish now to briefly discuss some of the connections and differences between our work and existing work in reinforcement learning.

The usual underlying structure in the reinforcement learning literature is that of an environment which changes as a result of an agent's action based on a particular probabilistic function. The agent's reward may be probabilistic as well.[16] In our notation, a Markov decision process (MDP) is a repeated game against Nature with complete information and strategic certainty, in which Nature's strategy depends probabilistically on the last action and state chosen[17]. Standard partially observable MDP (POMDP) can be described similarly by introducing a level of monitoring in between perfect and imperfect monitoring. In addition, bandit problems can be basically modeled as repeated games against Nature with a probabilistic structural assumption about Nature's strategy , but with strategic uncertainty about the values of the transition probabilities. For example, Nature's action can play the role of the state of a slot machine in a basic bandit problem. The main difference between the classical problem and the problem discussed in our setting is that the state of the slot machine may change in our setting in a totally unpredictable manner (e.g., the seed of the machine is manually changed at each iteration). This is not to say that by solving our learning problem we have solved the problem of reinforcement learning in MDP, in POMDP, or in bandit problems. In the later settings, our optimal strategy behave poorly relative to strategies obtained in the theory of reinforcement learning, that take the particular structure into account.

The non-Bayesian and qualitative setup call for optimality criteria which differ from the ones used in current work in reinforcement learning. Work in reinforcement learning discusses learning mechanisms that optimize the expected payoff in the long run. In a qualitative setting (as described above) long-run expected payoffs may not make much sense. Our optimality criteria expresses the need to obtain a desired behavior in most stages. One can easily construct examples where one of these approaches is favorite to the other one. Our emphasis is on obtaining the desired behavior in a relatively short run. Though, most analytical results in reinforcement learning have been concerned only with eventual convergence to desired behavior, some of the policies have been shown to be quite efficient in practice.

In addition to the previously mentioned differences between our work and work in reinforcement learning, we wish to emphasize that much work on POMDP uses information structures which are different from those discussed in this paper. Work on POMDP usually assumes that some observations about the current state may be available (following the presentation by Smallwood & Sondik, 1973), although observations about the previous state are discussed as well (Boutilier & Poole, 1996). Recall that in the case of perfect monitoring the previous environment state is revealed, and the immediate reward is revealed in both prefect and imperfect monitoring. It may be useful to consider also situations where some

16. The results presented in this paper can be extended to the case where there is some randomness in the reward obtained by the agents as well.

17. Likewise, stochastic games (Shapley, 1953) can be considered as repeated games against Nature with partial information about Nature's strategy. For that matter one should redefine the concept of state in such games. A state is a pair $(s, a)$, where $s$ is a state of the system and $a$ is an action of the opponent.

(partial) observations about the previous state or the current state are revealed from time to time. How this may be used in our setting is not completely clear, and may serve as a subject for future research.